\documentclass[letterpaper, 10 pt, conference]{ieeeconf}
\IEEEoverridecommandlockouts
\usepackage[utf8]{inputenc}
\usepackage{amsmath}
\usepackage{amssymb}
\usepackage{graphicx}
\usepackage{subcaption}
\usepackage{color}
\let\labelindent\relax
\usepackage[shortlabels]{enumitem}
\usepackage{hyperref}
\usepackage{bbm}
\usepackage{tabu}
\usepackage{booktabs}
\usepackage{multirow}
\usepackage{float}
\usepackage
[backend=bibtex,
bibstyle=ieee,
citestyle=numeric,
sortcites,
natbib=true,
doi=false,
isbn=false,
url=true,
hyperref=true,
sorting=nyt,
eprint=false,
maxbibnames=99]{biblatex}

\newcommand{\T}{\intercal}
\newcommand{\co}[1]{\texttt{#1}}

\usepackage{algorithm,algorithmicx}
\usepackage[noend]{algpseudocode}

\definecolor{Blue}{rgb}{0,0,1}
\definecolor{Orange}{rgb}{1,0.65,0}

\title{Integrating Open-World Shared Control in Immersive Avatars}
\author{Patrick Naughton$^{*1}$, \textit{Student Member, IEEE}, James Seungbum Nam$^{*2}$, \textit{Student Member, IEEE},\\Andrew Stratton$^1$, and Kris Hauser$^1$, \textit{Senior Member, IEEE}
\thanks{$^{1}$P. Naughton, A. Stratton and K. Hauser are with the Department of Computer Science, University of Illinois at Urbana-Champaign, IL, USA.
        {\tt\footnotesize \{pn10, ars21, kkhauser\}@illinois.edu}}%
\thanks{$^{2}$J. S. Nam is with the Department of Mechanical Science and Engineering, University of Illinois at Urbana-Champaign, IL, USA.
        {\tt\small sn29@illinois.edu}}%
\thanks{This work was supported by NSF Grant \#2025782.}
\thanks{*Equal contribution. Corresponding author listed first.}%
}
\begin{document}

\maketitle

\begin{abstract}

Teleoperated avatar robots allow people to transport their manipulation skills to environments that may be difficult or dangerous to work in. Current systems are able to give operators direct control of many components of the robot to immerse them in the remote environment, but operators still struggle to complete tasks as competently as they could in person. We present a framework for incorporating open-world shared control into avatar robots to combine the benefits of direct and shared control. This framework preserves the fluency of our avatar interface by minimizing obstructions to the operator's view and using the same interface for direct, shared, and fully autonomous control. In a human subjects study (N=19), we find that operators using this framework complete a range of tasks significantly more quickly and reliably than those that do not.

\end{abstract}

\section{Introduction}

Teleoperation allows humans to sense and act in remote locations that may be hazardous or difficult to access. Recently, several groups have developed robot avatars \cite{schwarz_nimbro_2021, marquescommodity, luo_team_2023, vanbotics} that provide immersive interfaces for operators to control an entire robot body and transport their presence to a remote location. These systems have proven that avatars enable novice operators to intuitively inspect, navigate, and manipulate the remote environment, but even state-of-the-art systems lag behind human proficiency~\cite{XPRIZESystemsPaper2023}.

\begin{figure}[htbp]
    \begin{subfigure}[t]{0.65\linewidth}
        \includegraphics[height=5.1cm]{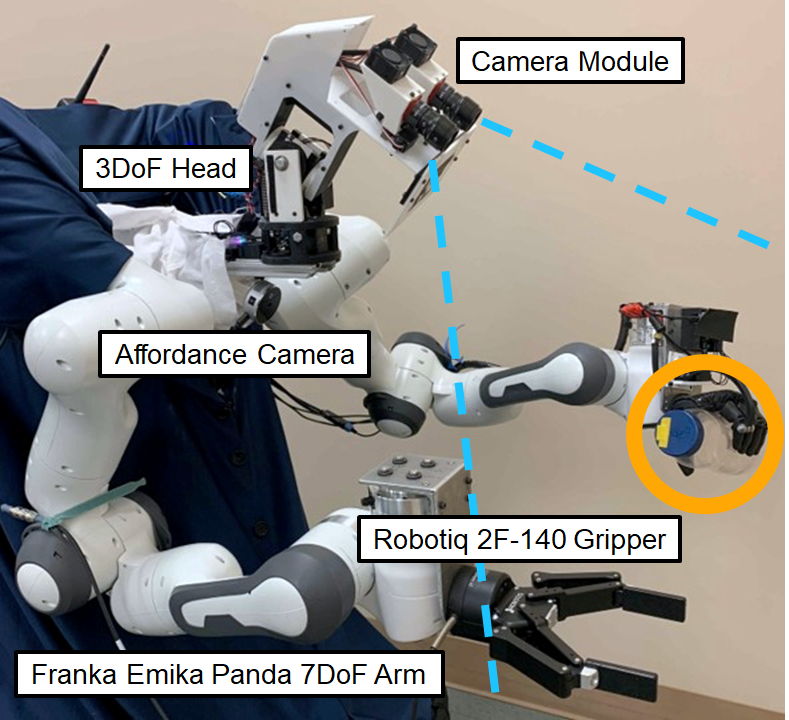}
        \caption{Robot holding jar in left gripper}
        \label{fig:robot_overview}
    \end{subfigure}
    \hfill
    \begin{subfigure}[t]{0.34\linewidth}
        \includegraphics[height=5.1cm]{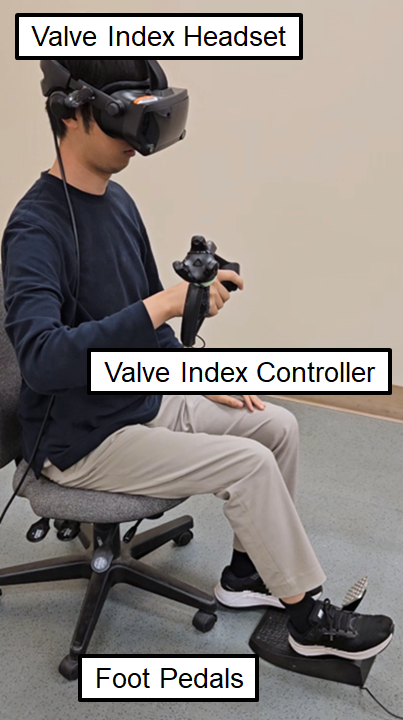}
        \caption{Operator interface}
        \label{fig:operator_overview}
    \end{subfigure}
    \begin{subfigure}[t]{0.49\linewidth}
        \centering
        \includegraphics[width=\linewidth]{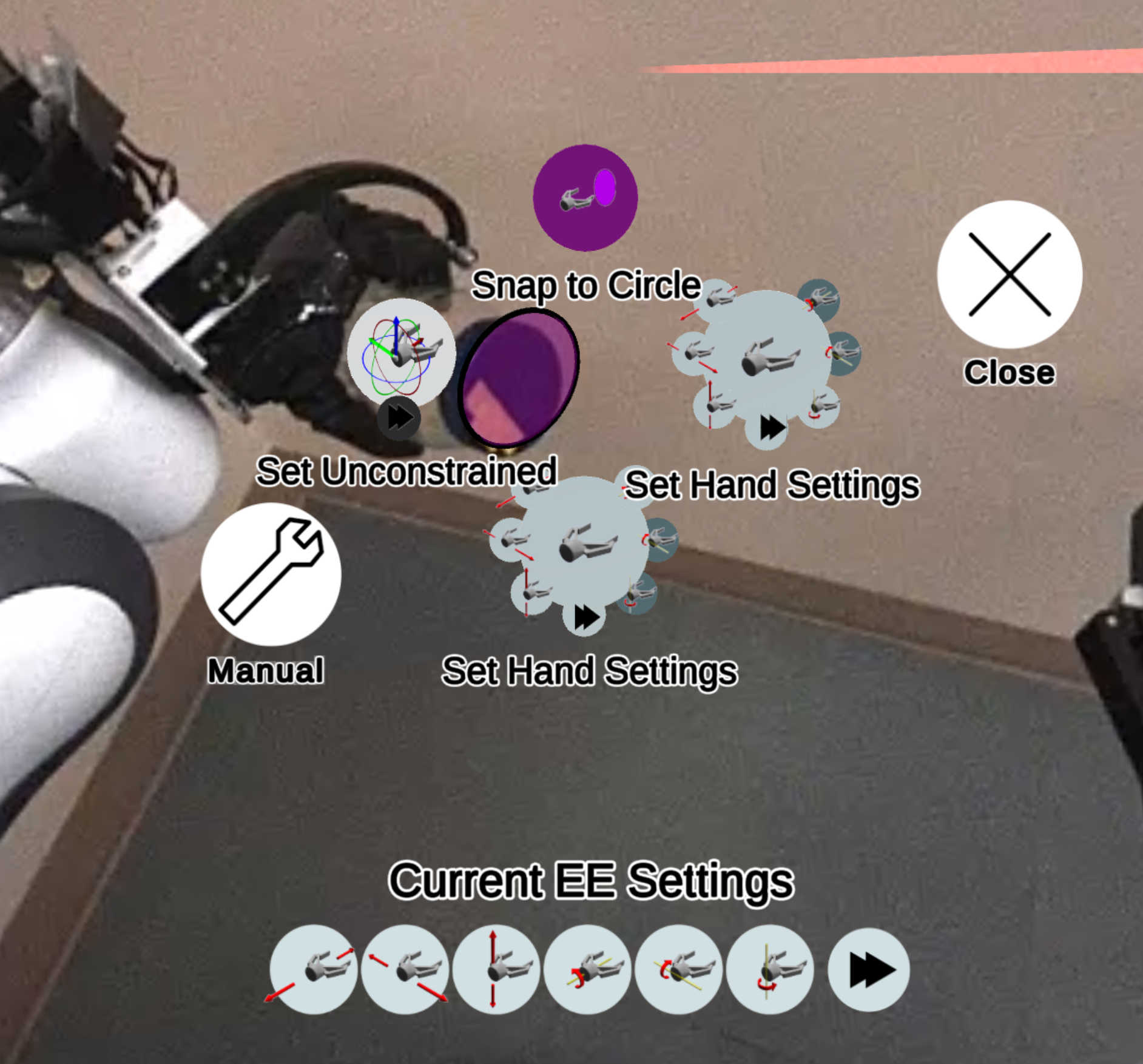}
        \caption{Operator's view with a predictive menu showing suggested actions}
        \label{fig:vr_overview}
    \end{subfigure}
    \hfill
    \begin{subfigure}[t]{0.49\linewidth}
        \centering
        \includegraphics[width=\linewidth]{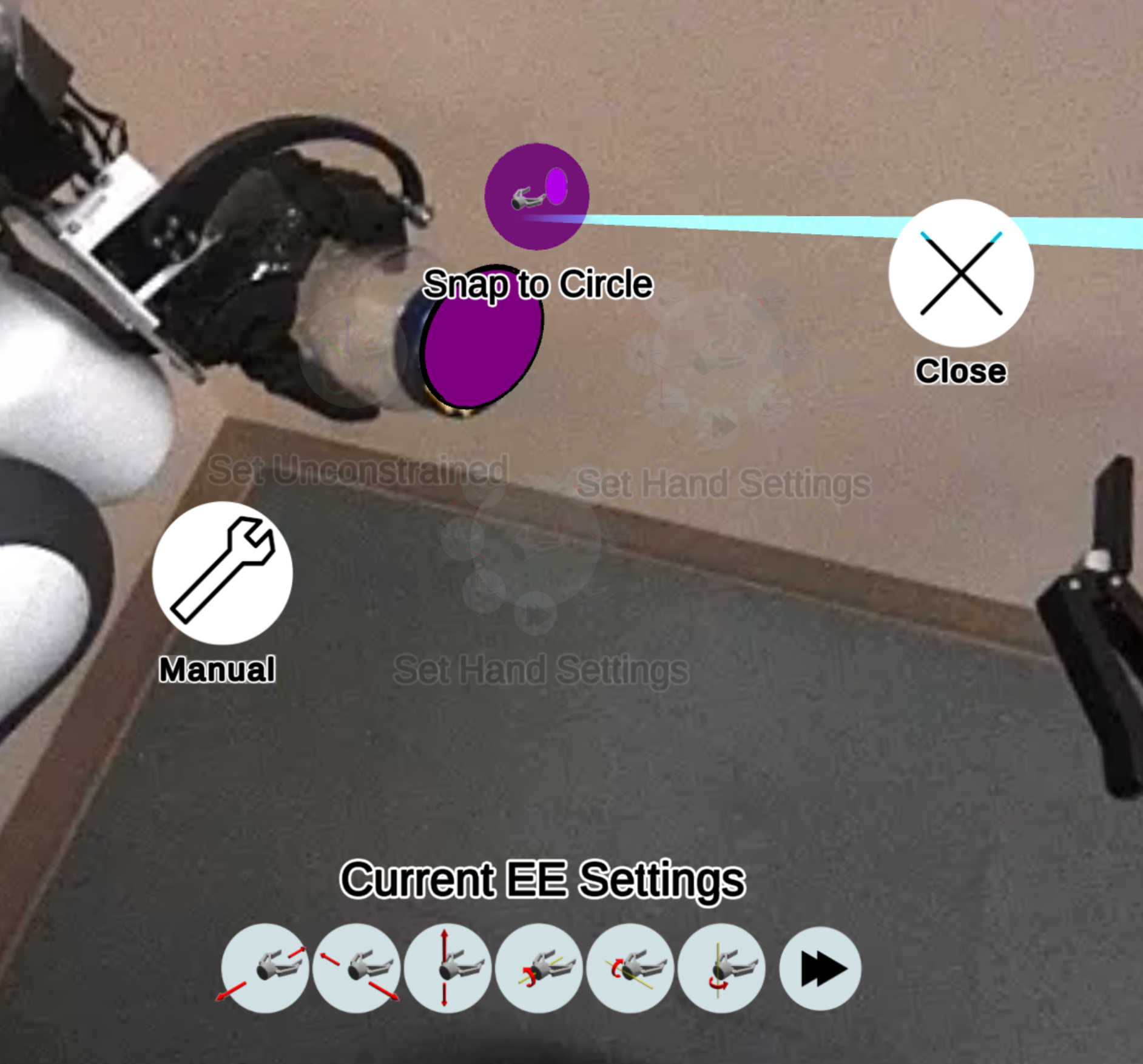}
        \caption{A ``laser pointer'' paradigm is used to select actions}
        \label{fig:vr_selecting_overview}
    \end{subfigure}
    \caption{An operator uses the Avatar robot to unscrew a jar using the immersive interface.  The predictive menu suggests possible assistive actions and shows corresponding affordances as augmented-reality objects (purple circle overlaying the jar lid). [Best viewed in color.]}
    \label{fig:overview}
    \vspace{-20pt}
\end{figure}

This skill gap has long been identified as an issue for teleoperation, and researchers have proposed many assistance schemes to mitigate it, including virtual fixtures~\cite{rosenberg_virtual_1993, bowyerActiveConstraintsVirtual2014, pruks_method_2022, huang_evaluation_2019}, mode switches~\cite{quereSharedControlTemplates2020}, and automated planning~\cite{leeperMethodsCollisionfreeArm2013, leeperStrategiesHumanintheloopRobotic2012, bustamante_cats_2022}. Assistance has been shown to help operators in structured lab settings, but several challenges remain before they can be deployed, such as ``open-world'' tasks (tasks where the number and/or types of objects in the robot's environment are not known a-priori)~\cite{young_review_2020}, predicting the operator's intent~\cite{li_classification_2023}, evaluating and managing the operator's trust~\cite{li_classification_2023}, and operator overload degrading the operator's fluency~\cite{fallonArchitectureOnlineAffordancebased}.  The open-world problem is particularly troublesome, since teleoperation is especially effective in leveraging human problem-solving and contextual understanding, but nearly all assistance methods are designed to work with predefined objects in semi-structured scenarios~\cite{bustamante_cats_2022, quereSharedControlTemplates2020, huang_evaluation_2019, wangTaskAutocorrectionImmersive}. Another major challenge is bridging assistance paradigms with the immersive paradigm. Existing avatars incorporate few assistive features~\cite{schwarz_robust_2023, luo_team_2023, AVATRINASystemsPaper}, whereas shared control literature typically considers non-immersive mouse and keyboard interfaces~\cite{leeperStrategiesHumanintheloopRobotic2012, pruks_method_2022}. The question of how to integrate these schemes introduces several design challenges, such as how to allow the operator to quickly switch between control modes and configure different types of assistance without occluding the view of the remote environment.

The contribution of this work is the design and evaluation of a framework to incorporate open-world shared control into immersive robot avatars.
To address the central design challenges highlighted above, we created an in-headset menu that allows the operator to launch and configure assistive actions using the same controllers they use to directly move the robot (\autoref{fig:overview}). We implement assistive actions based on geometric affordances that are agnostic to object identity, allowing them to work in a wide range of scenarios.  Affordances are rendered as augmented reality (AR) markers in the operator's immersive view when the user is configuring action targets.   We further enhance the fluency of this interface using an ``autocomplete'' predictive menu that predicts the operator's intent in the context of the current scene and history~\cite{naughton_structured_2022}. We incorporate this framework into an avatar system and evaluate novice users on long-form tasks that require many uses of the assistive actions. Human subjects testing $(N=19)$ verifies that our approach, with and without the predictive menu, increases task success rates and system usability, and decreases task completion times and operator workload over standard direct control interfaces while preserving the operator's self-reported sense of presence in the remote environment. 

\section{Related Work}

The recent ANA Avatar XPRIZE competition spurred rapid development of teleoperated avatar robots capable of transporting basic human manipulation skills to remote environments~\cite{XPRIZESystemsPaper2023}. 
As the competition emphasized immersion and presence, most teams made very little or no use of shared control, instead opting to give as much direct control to the operator as possible. This choice makes the systems open-world, immersive, and intuitive,
but users still struggle to perform tasks through the robot as proficiently as they would in-person~\cite{XPRIZESystemsPaper2023}. Shared control methods could hypothetically assist in operator proficiency while preserving desirable aspects of immersion, but mechanisms for achieving such integration are not well studied.

Operator assistance for non-immersive interfaces has received much attention in the literature. A significant line of work addresses reaching for an object~\cite{draganTeleoperationIntelligentCustomizable2013}, especially when the operator's interface has fewer DoFs than the robot~\cite{hauserRecognitionPredictionPlanning2013, javdani_shared_2015, quereSharedControlTemplates2020, Jeon-RSS-20}. In the avatar context, this is not normally a concern because the operator has access to high DoF input devices. Other research provides assistance for complex tasks but requires pre-programmed information about the environment and target objects~\cite{quereSharedControlTemplates2020, bustamante_cats_2022, huang_evaluation_2019}. For example, \cite{quereSharedControlTemplates2020} presents a system that can perform complicated tasks like opening a door, but key frames of reference for specific objects are labeled by hand, and the state-machines describing transitions between different phases of the tasks are pre-specified. Our work seeks to relax this requirement and provide assistance in an open-world where the semantic identities and number of objects encountered in the environment are not known ahead of time. We achieve this by using more generic types of assistance, detecting affordances at runtime rather than hand labelling them at design-time.

The work of Pruks and Ryu~\cite{pruks_method_2022} is most similar to our system. Similar to our system, their work uses off-the-shelf methods to segment the environment into geometric primitives and allows the operator to apply customizable virtual fixtures between features detected in the environment and features from the robot. However, they use a screen-and-mouse interface to specify virtual fixtures and a separate haptic device to input low-level motion commands, requiring the operator to switch between two input devices. In contrast, our system uses a consistent input interface for both specifying virtual fixtures and providing low-level commands. Our system also provides an immersive interface via a virtual reality headset, rather than a standard screen interface. Finally, we also present a framework for incorporating predictive assistance into our system, which ~\cite{pruks_method_2022} did not consider.

\section{Interface Design}

\begin{figure}[tbp]
    \centering
    \includegraphics[width=\linewidth]{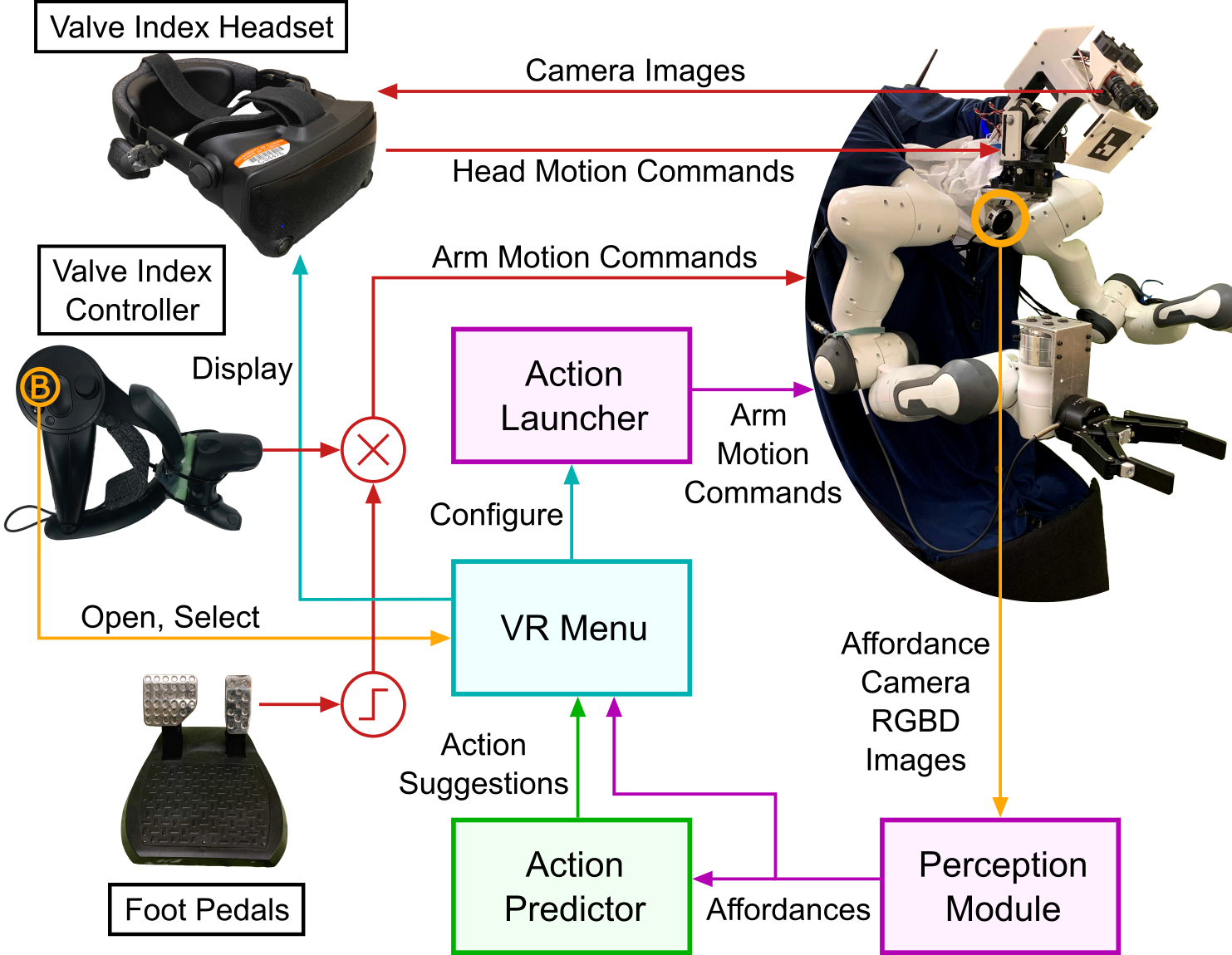}
    \caption{System diagram showing how different interface elements control the robot. Operators use their own head and hand to control the robot's head and hand, and use a button on their hand controller to interact with the assistive menu. The Perception Module detects affordances in the environment to display possible assistive actions to the operator. [Best viewed in color.]}
    \label{fig:system_diagram}
    \vspace{-15pt}
\end{figure}

\begin{figure*}[t]
    \centering
    \includegraphics[width=0.86\linewidth]{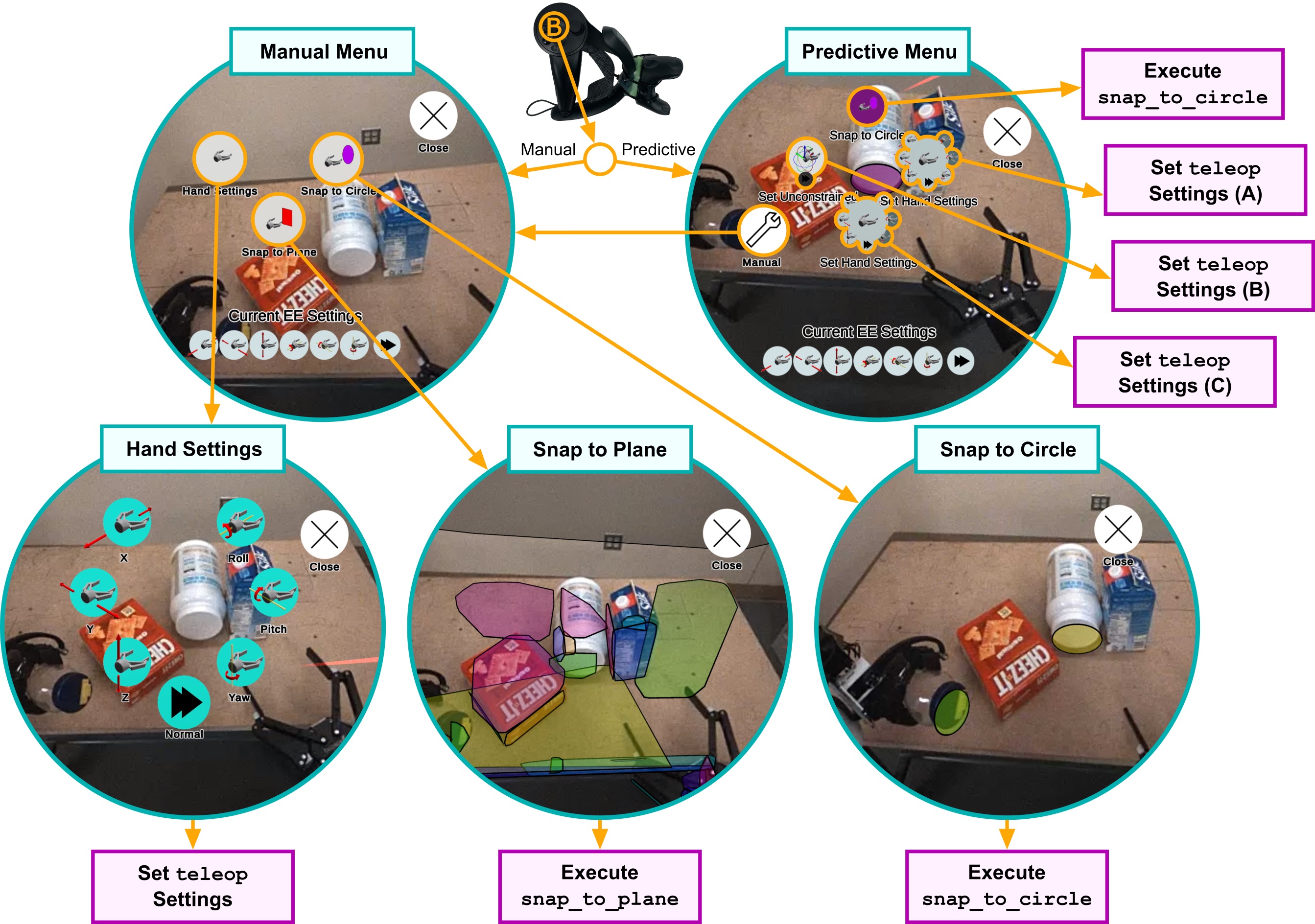}
    \caption{Flow diagram showing how different menus are accessed. Depending on which interface type is being used, the B button will show the operator different interfaces: in manual mode, this button will directly show the manual menu, while in predictive mode, it will show the predictive menu. In the predictive menu shown here, each \co{teleop} icon gives the operator the option to choose a different set of constraints. Orange emphasis is added to highlight certain icons, and is not present in the actual menu. [Best viewed in color.]}
    \label{fig:menu_flow}
    \vspace{-15pt}
\end{figure*}

Suppose that an avatar robot has a library of assistive actions available which may include shared control and semi-autonomous actions. The key design question is {\em how to let the operator access and configure assistive actions without breaking immersion and maintaining or enhancing fluency?}  Our approach is designed to satisfy the following objectives:
\begin{itemize}
    \item O1. The operator must be able to quickly switch between direct, shared, and autonomous control modes.
    \item O2. The same control and feedback interfaces must be used for each level of control.
    \item O3. The operator should be able to see as much of the remote environment as possible even when configuring assistive actions.
    \item O4. The robot should determine which target objects for actions are available dynamically, i.e., from open-world perception applied to the robot's current context.
    \item O5. The interface should have a limited number of displays and widgets to minimize operator overload and facilitate faster learning.
\end{itemize}
We build our work on the TRINA avatar system \cite{AVATRINASystemsPaper}, in which the robot is comprised of two Franka Emika Panda arms, a Robotiq 2F-140 parallel-jaw gripper, an anthropomorphic Psyonic Ability Hand, a Waypoint Vector omnidirectional wheeled base, and a custom-built three DoF neck and head assembly. A human operator controls TRINA using a virtual reality (VR) head-mounted display (HMD) that shows the view of TRINA's environment from stereo head cameras. They control the robot's head directly via HMD motion and use VR controllers to move the arms. The operator station is connected to the Internet via Ethernet and the robot is connected via WiFi or an Ethernet tether.

\autoref{fig:system_diagram} illustrates the major components of the proposed interface. Specifically, to satisfy O1 and O2, action selection functions are triggered with a single controller button. To satisfy O3, an unobtrusive VR Menu with a hierarchical pie system is overlaid atop the camera feed to configure and launch actions. For O4, the Perception Module continually recognizes geometric affordances in the robot's environment, which are rendered as selectable AR objects.  For O5, we incorporate a machine learning-based Action Predictor to generate a Predictive Menu trained on expert demonstrations.

\subsection{Direct Teleoperation (DT)}

The default control mode is the direct teleoperation scheme described in~\cite{AVATRINASystemsPaper}. To simplify novice operator training, in our experiments, we only activate the robot's right arm, parallel-jaw gripper, and head. The operator wears a VR HMD and the robot's head tracks the operator's head orientation. 
The operator uses a clutched system to control the arm: while holding down a foot pedal, the operator moves a VR controller, shown in \autoref{fig:system_diagram}, to move the robot's hand target. This motion is computed relative to the controller's pose when the operator first presses the pedal. A lower-level controller then attempts to reach this target. 
The operator can also velocity-control the parallel-jaw gripper using a joystick on the controller, pushing it right to inch the gripper closed, and left to inch it open.

The robot estimates the net force applied to its end effector to provide force feedback via two modalities: First, the controller vibrates with an intensity proportional to the estimated force magnitude (clipped between 10 and 30 N). 
Second, a virtual red hemisphere around the operator's controller shows the direction of the applied force, and becomes more opaque as the magnitude of the force increases.

\subsection{Manual Menu (MM)}

Using the direct teleoperation interface alone, operators can achieve some manipulation tasks~\cite{AVATRINASystemsPaper}, but complicated tasks, such as writing, are still quite difficult. To aid the operator, we created an interface to allow them to execute assistive actions. Guided by previous research~\cite{komerska_study_2004}, we designed a hierarchical pie menu fixed to the operator's head,
shown in \autoref{fig:menu_flow}. By making the menu hierarchical, we minimize the number of simultaneously displayed icons to keep the operator's view of the remote environment unobstructed. 
The operator interacts with the menu using a ``laser pointer'' emanating from their controller to point at different icons, and clicks the B button on their controller to select them. The operator can bring up this menu by clicking the B button at any time and can close it by selecting the ``Close'' icon. This menu design allows the operator to configure the menu using the same interface they use to provide low-level commands to the robot, eliminating any need to switch between interfaces during operation.
Clicking other icons gives the operator access to different submenus.

The ``Hand Settings'' submenu allows the operator to edit constraints and the sensitivity mode of the arm by selecting any of the icons to toggle their state.
The ``Snap to Plane'' and ``Snap to Circle'' submenus display the most recently detected affordances of each type, shown in \autoref{fig:menu_flow}. Each affordance is rendered as an AR object in the virtual world, displayed so that it appears aligned with the object it was detected from, with a random hue at 30\% opacity. By performing this alignment, the menu leaves the operator's view essentially unobstructed, integrating information about affordances with the operator's existing view of the environment.
When the operator hovers over an affordance with their laser pointer, that affordance becomes opaque. Selecting an affordance will send it to the robot, which will then execute the corresponding action.


Whenever the operator selects an action, ``Executing Action'' followed by ``Action Succeeded'' or ``Action Failed'' is displayed depending on its status. If an action fails, the arm maintains the position it had when the failure occurred. The operator can also cancel actions by pressing their foot pedal, which gives them direct control over the arm as usual.

\subsection{Predictive Menu (PM)}

While the manual menu provides access to all possible actions, it can be overwhelming and slow, especially for novice users. 
To alleviate this, we designed a third interface that uses an action predictor, described in \autoref{sec:intent}, to predict the operator's intent and present them with a reduced menu that only includes the four most likely actions. If the operator's desired action is not in this set, they can still access the manual menu as a fallback. With this menu, when the operator clicks B, the top four actions are shown instead of the manual menu, as shown in \autoref{fig:vr_overview} and \autoref{fig:menu_flow}. 
Whenever the operator hovers over an icon corresponding to an action, all other icons (and affordances) dim to 10\% opacity. Selecting any icon closes the menu and sends the action to the robot which then executes it.

We assume that the robot is the only agent in the scene and that all manipulations are quasistatic. As a result, the state of the world only changes when the robot is executing an action. Therefore, we design the robot to run the action predictor to produce the next set of suggestions when it first starts up, and after any action is completed.
While these assumptions do not strictly hold in all experiments, they are good enough approximations to produce accurate predictions while not having to compute new predictions in every frame.

\section{Assistive Actions}\label{sec:assistance}

We implemented three kinds of assistive actions: constrained teleoperation, snapping to planes, and snapping to circles. The use of geometric affordances to provide assistance allows the use of these actions in an open-world context, where the semantic meaning of objects in the environment is unknown. The constrained teleoperation and plane snapping actions were previously described in~\cite{naughton_structured_2022}, and so are only briefly covered here.

The constrained teleoperation action, \co{teleop(sens, x, y, z, roll, pitch, yaw)} accepts 7 Boolean parameters modifying the operator's direct control of the arm. During this action, the operator controls the gripper's target pose by moving a VR controller with their own arm. When the \co{sens} parameter is \co{true}, the arm's end-effector motion is isotropically scaled to $0.25$ of the operator's input motion to enable precise manipulation. The remaining parameters toggle constraints on the end-effector motion, activating guidance virtual fixtures to simplify operation~\cite{bowyerActiveConstraintsVirtual2014}.

\begin{figure}[tbp]
    \centering
    \includegraphics[height=5.3cm]{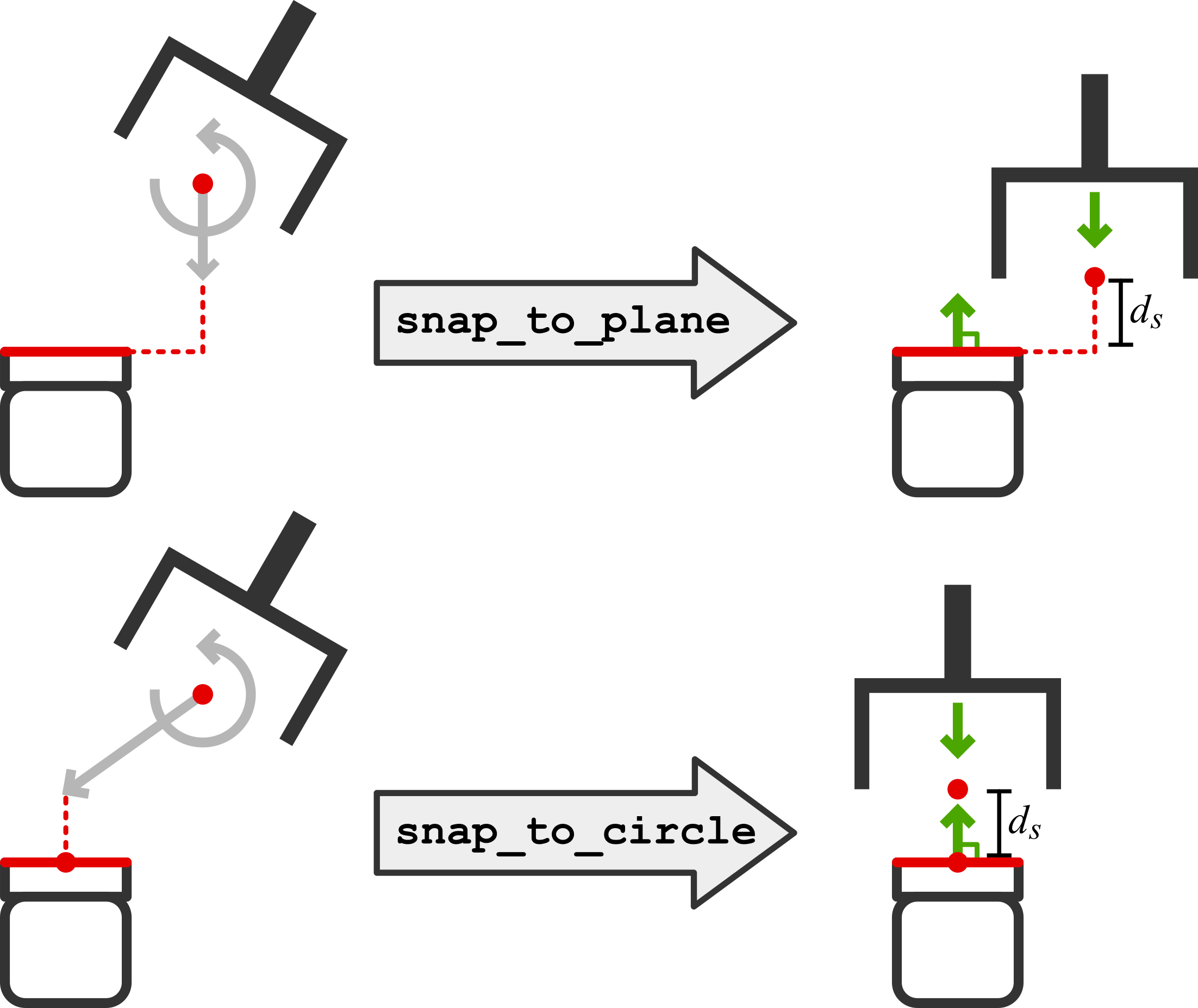}
    \caption{2D illustration of the \co{snap\_to\_plane} and \co{snap\_to\_circle} actions. Both align TRINA's gripper with the normal of the selected affordance, but \co{snap\_to\_circle} centers the gripper on the circle while \co{snap\_to\_plane} only moves it closer to the plane. Here we used $d_s = 0.15$~m. [Best viewed in color.]}
    \label{fig:snap_actions}
    \vspace{-20pt}
\end{figure}

The plane snapping action, \co{snap\_to\_plane(p)} accepts a plane detected from a point-cloud of the environment by a clustering method~\cite{fengFastPlaneExtraction2014}. This point-cloud is sensed by the ``affordance camera'' shown in \autoref{fig:system_diagram}, an Intel RealSense L515 mounted below the robot's neck, pointed at the center of the robot's workspace. The plane extraction algorithm updates the set of detected planes once every 5 seconds. This action aligns the forward direction of the gripper with the normal of the detected plane and moves it so that its tool tip is $d_s$~m away from the plane to prepare the operator to perform manipulation on or near the plane's surface. For the tasks considered here we found $d_s = 0.15$~m to work well. \autoref{fig:snap_actions} illustrates this process in 2D. The robot uses a sampling-based planner to find a path to reach this target or reports that no path was found after 10~s.

Lastly, the \co{snap\_to\_circle(c)} action accepts a circle detected from the environment, aligns the gripper's forward direction with the circle's axis, and centers the gripper on the circle to prepare the operator to perform rotating manipulations about the circle's axis. Our system detects circles from RGBD images from the affordance camera once every 5 seconds. The system segments the RGB image using the Segment Anything Model (SAM)~\cite{kirillov2023segany} and converts the RGBD image into a point-cloud.
For each image mask, the corresponding points are selected, and the plane supported by the most points is found.
The inliers of this plane are computed as the points in the mask within $d_\text{in} = 5\text{ mm}$ of the plane and projected to the plane. The convex hull of these projected points is found and the circle is discarded if this hull's ``circularity'' ($\frac{4\pi\cdot \text{Area}}{\text{Perimeter}^2}$ \cite{opencv_library}) is below $c_\text{min} = 0.9$. The minimum enclosing circle of the hull is computed and circles with radii greater than $r_\text{max} = 7\text{ cm}$ are discarded. To remove duplicates, this candidate circle is compared against previously detected circles. Circles are considered similar if the masks from which they were detected overlap, their centers are within $\Delta_\text{c} = 5\text{ cm}$, and their radii are within $\Delta_\text{rad} = 1\text{ cm}$. Among similar circles, the one with the largest ratio of inliers to points in the mask is kept. Once a circle has been selected, the robot computes a target end-effector pose in the same manner as the \co{snap\_to\_plane} action, additionally moving the target so that the projection of the tool tip to the plane of the circle coincides with the circle's center. \autoref{fig:snap_actions} demonstrates this action in 2D.

\section{Intent Prediction}\label{sec:intent}

To populate the predictive menu, we require an action predictor that can predict multiple likely actions. Additionally, since the set of affordances is not known until runtime, the predictor must be open-world, i.e. able to predict over an open set of objects. We employ the structured prediction method of \cite{naughton_structured_2022} as it was found to have strong performance in open-world scenarios on similar tasks. 

Actions are defined by a type and a collection of parameters, $\overline{\psi}$, which may be different for each action type. We limit the set of $n$ types \textit{a priori} and dynamically detect the set of feasible parameters for each type, corresponding to detected affordances.
To predict an action given the robot's current context vector, $x$, the method uses $n$ parameter scoring neural networks, $\{G^{(i)}(x, \overline{\psi})\}_{i=1}^n$, and an action network, $A(x)$.
$A(x)$ produces an $n$-dimensional output vector with each element representing the overall score for an action type. Each $G^{(i)}(x, \overline{\psi})$ predicts a scalar score for parameter collections of a particular action type. 
To score a complete action, the appropriate scores are summed, $s=e_i^\T A(x) + G^{(i)}(x, \overline{\psi})$, where $e_i$ is the $i$th standard basis vector.

To train and evaluate our predictor, three expert operators (paper authors) collected a dataset of 150 action sequences across three different tasks: unscrewing a jar lid, writing ``IML'' on a whiteboard, and plugging a cord into an electrical socket. 
Each sequence was collected in a highly cluttered environment that contained many different distractor objects with varied compositions and arrangements. The specific target objects used were also modified (for example, varying which jars were used).
The scoring function was trained using a maximum margin loss function to output high scores for actions observed in the demonstrations~\cite{naughton_structured_2022}.

\section{Experiments}\label{sec:human_subjects_experiments}

\begin{table*}[!ht]
    \centering
    \caption{Differences between each interface across all tasks. ${}^{*}$, ${}^{**}$, and ${}^{***}$ denote $p\leq 0.05$, $p\leq 0.01$, and $p\leq 0.001$ respectively.}
    \label{tab:overall_results}
    \begin{tabu}{X[1.5] X X[r] X[r] X[r] X[r] X[r]}
        \toprule
        & Condition & Success (\%) $(\uparrow)$ & Time (s) $(\downarrow)$ & Usability $(\uparrow)$ & Workload $(\downarrow)$ & Presence $(\uparrow)$ \\
        \midrule
        \multirow{3}{*}{Avg $\pm$ Std} & DT & 42.7 $\pm$ 30.5 & 756 $\pm$ 179 & 4.32 $\pm$ 1.04 & 5.29 $\pm$ 1.14 & 4.53 $\pm$ 1.68 \\
        & MM & 68.5 $\pm$ 28.9 & 672 $\pm$ 183 & \textbf{5.17 $\pm$ 0.56} & 4.21 $\pm$ 1.26 & \textbf{5.00 $\pm$ 1.29} \\
        & PM & \textbf{75.8 $\pm$ 24.2} & \textbf{650 $\pm$ 152} & 5.01 $\pm$ 0.74 & \textbf{3.90 $\pm$ 1.13} & 4.74 $\pm$ 1.33 \\
        \midrule
        Friedman W-Score & & 0.4014 & 0.2696 & 0.2647 & 0.3836 & 0.0269 \\
        \midrule
        Friedman p-value & & ${}^{***}$0.0005 & ${}^{**}$0.0060 & ${}^{**}$0.0065 & ${}^{***}$0.0007 & 0.6004 \\
        \midrule
        \multirow{3}{*}{Post-hoc p-value} & DT vs. MM & ${}^{**}$0.0066 & 0.0611 & ${}^{**}$0.0015 & ${}^{**}$0.0053 & 0.1308 \\
        & DT vs. PM &  ${}^{***}$0.0004 & ${}^{*}$0.0115 & ${}^{**}$0.0061 & ${}^{***}$0.0004 & 0.5202 \\
        & MM vs. PM & 0.4844 & 0.5412 & 0.2882 & 0.2958 & 0.3543 \\
        \bottomrule
    \end{tabu}
    \vspace{-18pt}
\end{table*}

Human subjects studies were conducted to evaluate differences between the DT, MM, and PM interfaces. All procedures were reviewed and approved by the UIUC IRB on Feb. 20, 2023. We formulated the following \textit{a priori} hypotheses about the system:
\begin{itemize}
    \item \textbf{H1}: There is a difference in the proportion of tasks operators complete when using each interface.
    \item \textbf{H2}: There is a difference in the operators' total task completion times when using each interface.
    \item \textbf{H3}: There is a difference in the operator's sense of presence when using each interface.
\end{itemize}

\begin{figure}[tbp]
    \begin{subfigure}[t]{0.31\linewidth}
        \centering
        \includegraphics[width=\linewidth]{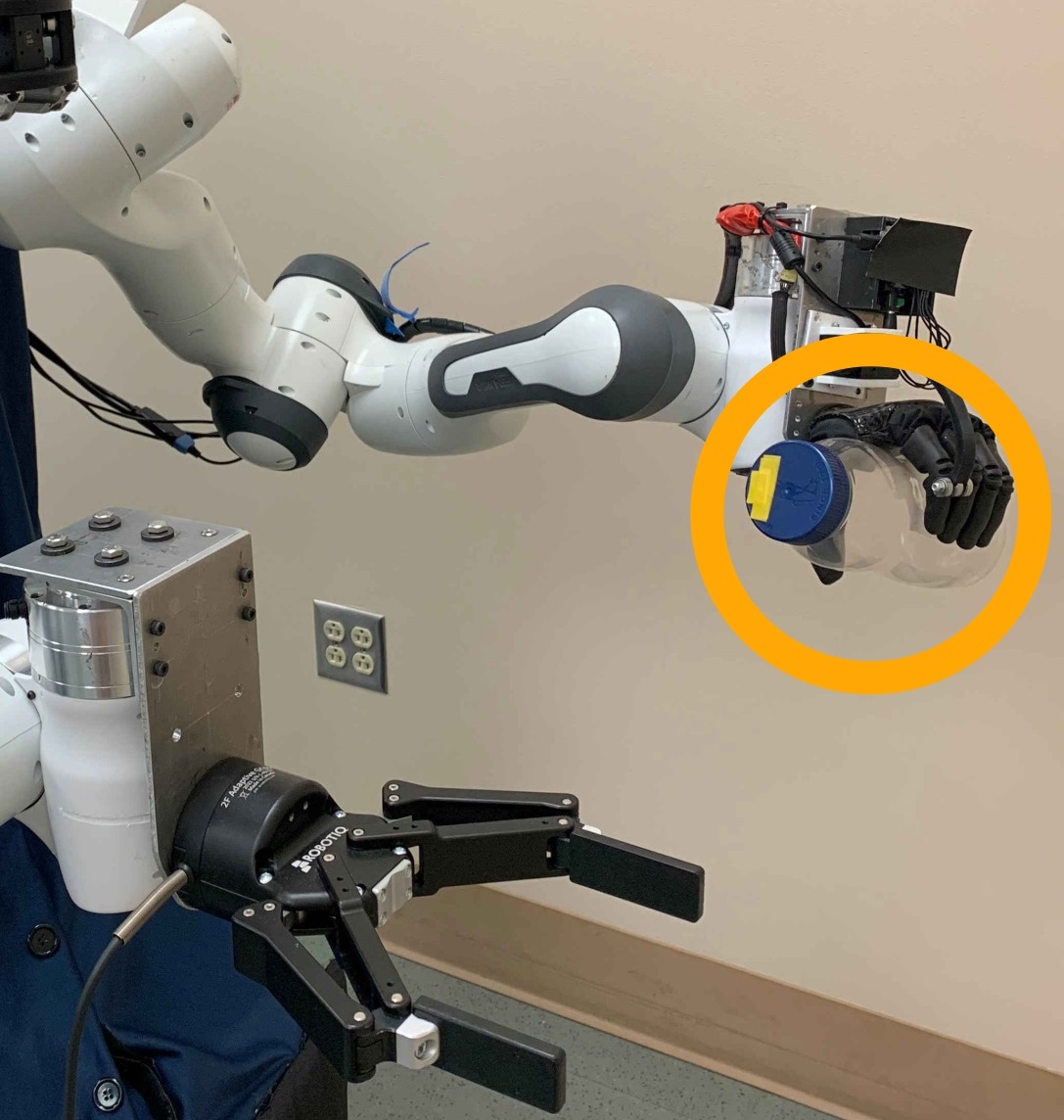}
        \caption{Jar}
        \label{fig:jar_test}
    \end{subfigure}
    \hfill
    \begin{subfigure}[t]{0.31\linewidth}
        \centering
        \includegraphics[width=\linewidth]{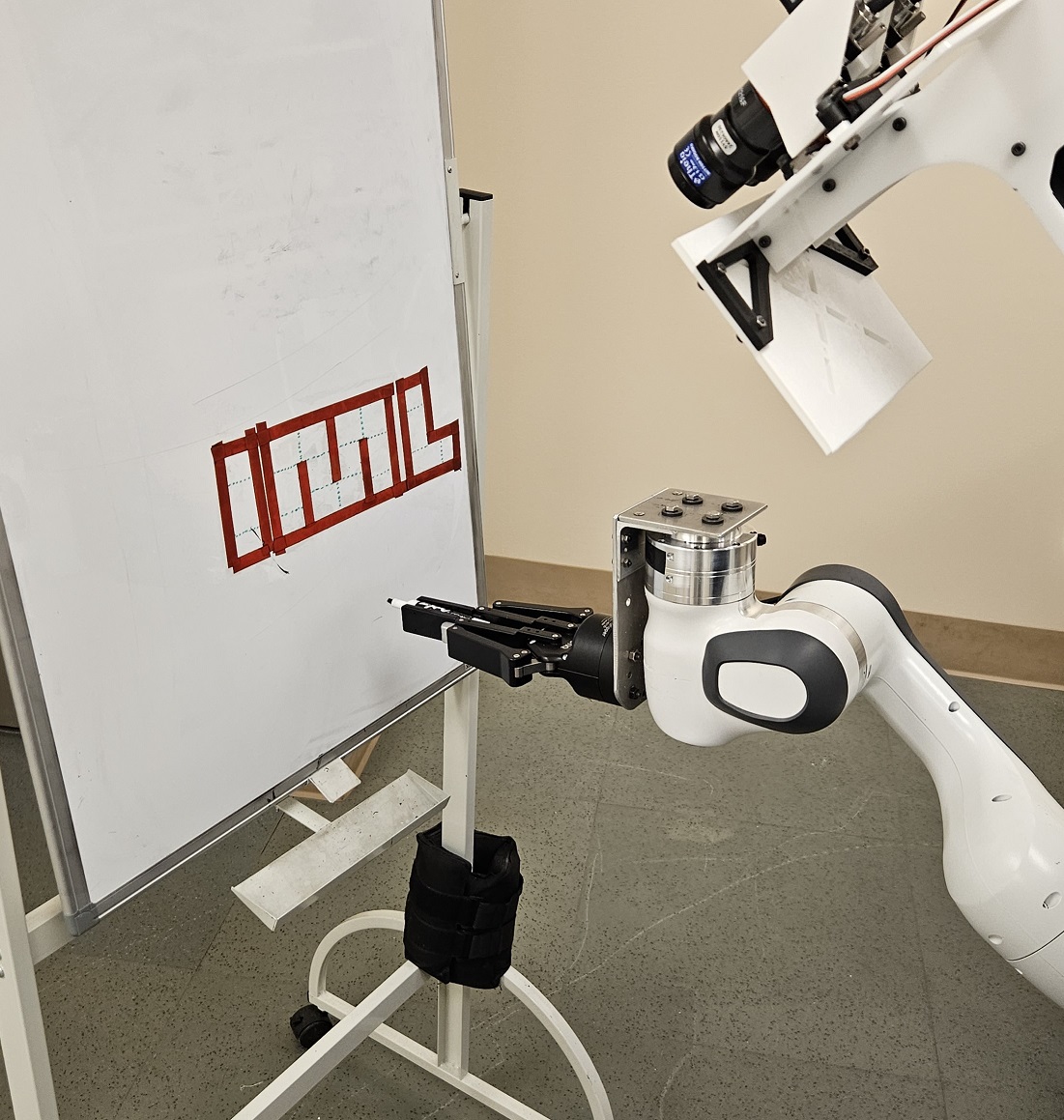}
        \caption{Whiteboard}
        \label{fig:whiteboard_test}
    \end{subfigure}
    \hfill
    \begin{subfigure}[t]{0.31\linewidth}
        \centering
        \includegraphics[width=\linewidth]{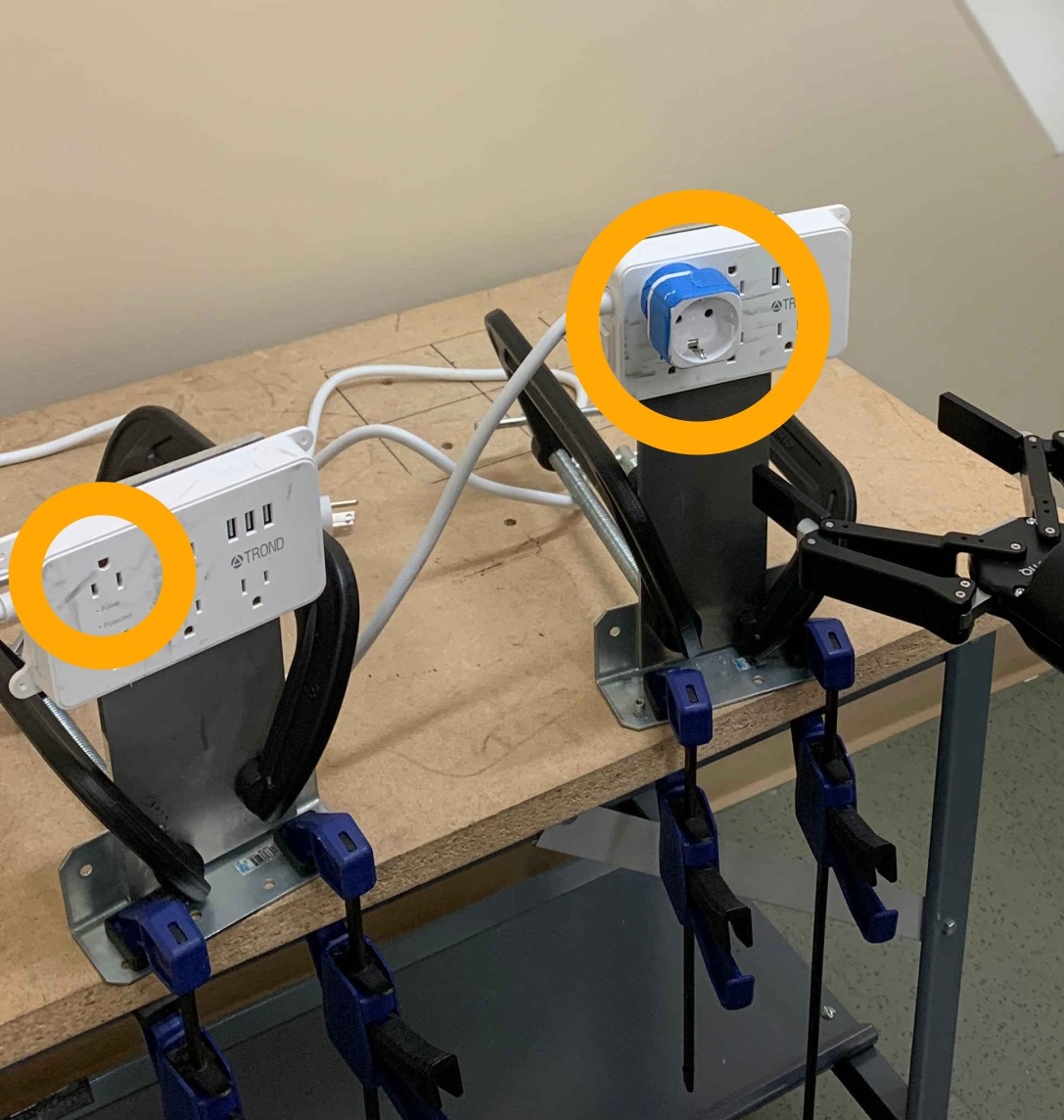}
        \caption{Plug}
        \label{fig:plug_test}
    \end{subfigure}
    \caption{The three testing tasks. Target objects are highlighted with orange circles. [Best viewed in color.]}
    \label{fig:test_tasks}
    \vspace{-20pt}
\end{figure}

To test our hypotheses, we designed a human subjects study to test novices' use of each interface. We considered three tasks: unscrewing a jar lid held in TRINA's left hand, writing ``IML'' on a whiteboard, and plugging in an electrical plug. Setups for these tasks are shown in \autoref{fig:test_tasks}.  The predictor was trained on expert demonstrations of the same tasks. These tasks were chosen to be representative of multi-stage tasks in which assistance is useful but solution strategies are somewhat flexible; novice strategies can differ significantly from one another and the expert demonstrations. 

We recruited 20 student participants from the University of Illinois at Urbana-Champaign campus, 19 of whom completed the entire procedure. One participant requested to end the experiment during training due to nausea. Of the 19 participants, 11 were male, 7 were female, and one preferred not to say. Subjects were of age 19--32 (mean: 24) and self-reported their familiarity with robotics and controlling robots on average as 5.4 and 4.4 on a 7-point Likert scale \cite{sarantakos2017social} respectively. None of the subjects had used TRINA before.


\subsubsection{Basic Training}

Subjects were trained to use the direct teleoperation interface
and were introduced to several possible fault states. For example, if excessive force was applied to the arm, the subject would momentarily lose control of it. Subjects were given suggestions about how to resolve each of these faults. The assistive functionalities were demonstrated using the manual (MM) and predictive (PM) menus.

\subsubsection{Task Introduction}

Subjects were shown the three testing tasks and completed the tasks in-person to familiarize themselves with the specific features of the target objects.
A researcher explained how task completion would be graded, and that subjects should try to complete tasks as quickly as possible with 5~min at most for each task. For the jar, the task was completed when the lid no longer was touching the jar body. For the whiteboard, the required writing was split into 19 segments and credit was given for each completed segment.
For the plug, the task was completed when the subject had fully inserted the plug into the target socket.

\subsubsection{Training Tasks}

Subjects were coached through using the MM and PM on two training tasks which demonstrated each of the assistive actions in context. In the first task, a researcher handed TRINA a capped Expo marker, and the subject had to use TRINA to insert the tip of the marker into a square hole.
Subjects were told to snap to the plane of the hole and turn off all rotational DoFs before inserting the marker into the hole. In the second task, subjects had to grasp and turn a dial 
for three full rotations. They were instructed to first snap to the circle of the dial, disable all but the \co{x} and \co{roll} DoFs to grasp the dial, and finally have only \co{roll} enabled to turn the dial.


\subsubsection{Testing Procedure}

On average, training took $\sim$90~min. After training, the order of conditions (DT, MM, and PM) was randomized.
For each condition, subjects completed the tasks in the order of jar, whiteboard, then plug. Subjects were given 3 and 1~min remaining warnings.
To minimize variance between the subjects, the placement of the target objects in the scene was kept consistent, and there were no distractor objects. Additionally, the jar and plug were modified to make the tasks slightly easier for novices: bright tape was added to the lid of the jar,
and a socket adapter was used as the plug instead of an electrical cord. Blue tape was also added to the adapter to make it easier to see. 
After attempting all of the tasks in a given condition, subjects filled out a questionnaire about their experience, measuring the system's usability \cite{brooke1996sus}, the subject's workload \cite{hart_development_1988}, and their self-reported feeling of presence in the remote environment. All questions were rated on a 7-point Likert scale. Subjects would then immediately proceed to the next condition.

\section{Results and Discussion}

Subject performance was measured by the proportion of tasks completed and the time taken. Success metrics are computed as $(\text{Did jar} + \text{Segments completed} / 19 + \text{Did plug}) / 3$. If a subject failed a task early, their time was recorded as the maximum time. We ran a Shapiro--Wilk test \cite{shapiro_analysis_1965} on the performance metrics for each condition and found significant deviations from normality. To test \textbf{H1}, \textbf{H2}, and \textbf{H3} we ran separate Friedman tests \cite{sarantakos2017social} on the subjects' success rates, completion times, and reported senses of presence, which revealed significant differences between the conditions for success rates ($p=0.0005$) and completion times ($p=0.0060$), but not for senses of presence ($p=0.6004$). Post-hoc pairwise two-sided Wilcoxon-signed-rank testing \cite{sarantakos2017social} found a significant increase in success rate for DT vs. MM ($M = 25.9\%, SD = 33.6\%, p=0.0066$) and DT vs. PM ($M = 33.1\%, SD = 26.6\%, p=0.0004$), and a decrease in completion time for DT vs. PM ($M = 105\text{ s}, SD = 170\text{ s}, p=0.0115$). 
\autoref{tab:overall_results} shows these results and includes results of exploratory analysis performed on other subjective measures, indicating that the presented interfaces also improve usability and workload.

These results provide support for \textbf{H1} and \textbf{H2}, indicating that the presented system can significantly improve novice operators' ability to perform several tasks quickly and accurately. 
We also found that the predictive menu generally has a larger impact on both objective and subjective metrics than the manual menu, despite its relatively low accuracy of 60.\% on novice actions. We expect this impact to further increase as the number of possible actions and the accuracy of the predictor rise.
The lack of support for \textbf{H3} suggests that this menu system preserves the operator's sense of presence despite introducing non-physical visual elements; in fact, both MM and PM received higher average presence scores than DT. We attribute this to the minimally invasive nature of the hierarchical pie menu and affordances registered to the remote environment. We further found that both the MM and PM interfaces tend to increase the system's usability and decrease the operator's workload. Users can easily understand how to interact with both kinds of menus and use them to decrease the required cognitive effort to complete manipulation tasks.

Our results show that contrary to conventional wisdom, designers of avatar robots need not choose between an immersive interface and using shared control: it is possible to achieve both in a single system. When integrating these two control paradigms, we suggest designers follow the philosophy presented here. For example, for shared control actions that reference the robot's environment, directly overlaying visual elements corresponding to those actions onto the operator's existing view lets the operator launch those actions while still focusing on their desired task. The manual menu presented here keeps the number of simultaneously presented icons low using a hierarchy, and this can be further improved for systems with large numbers of actions by using a predictive menu.



\section{Conclusion}

Our unified interface demonstrates a route for robot avatars to harness the ``best of both worlds'' between immersive teleoperation and assistive actions.  Our interface gives avatar operators intuitive access to assistive actions with dynamic affordance detection and AR overlays in an unobtrusive menu, and experiments showed that our approach improves operator fluency on three multi-step tasks without degrading immersion. 
In future work, we would like to expand the set of assistive actions to include automatic grasping and tool-centric shared control. We also wish to study how the interface affects operator performance in longer-form tasks, and to develop action predictors that adapt to individual operators online.

\printbibliography

\end{document}